\documentclass[10pt, a4paper]{article}
\usepackage[top=2cm,left=3cm,footskip=2cm,marginparwidth=2.2cm]{geometry}

\usepackage{amsmath, amsfonts}
\usepackage{graphicx}
\usepackage{multirow}
\usepackage{harvard}
\usepackage{lmodern}
\usepackage{url}
\usepackage[utf8]{inputenc}
\usepackage{nameref,hyperref}
\usepackage{microtype}
\usepackage{float, subfig}
\DisableLigatures[f]{encoding = *, family = * }

\usepackage[font=small]{caption}

\providecommand{\keywords}[1]
{
  \small	
  \textbf{\textit{Keywords---}} #1
}

\hypersetup{pdfauthor={WillNash}, pdftitle={Datasetpaper}}

\graphicspath{ {images/} }
\title{Quantity beats quality for semantic segmentation of corrosion in images}
\author{Will Nash\thanks{W.~Nash (will.nash@monash.edu) PhD student at Monash University, Australia}\and Tom Drummond\thanks{T.~Drummond is a Professor of Electrical and Computer Systems Engineering at Monash University.}\and Nick Birbilis\thanks{N.~Birbilis is a Professor, and Head of Materials Science and Engineering at Monash University} }

\begin{document}

\maketitle

\begin{abstract}
Dataset creation is typically one of the first steps when applying Artificial Intelligence methods to a new task; and the real world performance of models hinges on the quality and quantity of data available. Producing an image dataset for semantic segmentation is resource intensive, particularly for specialist subjects where class segmentation is not able to be effectively farmed out. The benefit of producing a large, but poorly labelled, dataset versus a small, expertly segmented dataset for semantic segmentation is an open question. Here we show that a large, noisy dataset outperforms a small, expertly segmented dataset for training a Fully Convolutional Network model for semantic segmentation of corrosion in images. A large dataset of 250 images with segmentations labelled by undergraduates and a second dataset of just 10 images, with segmentations labelled by subject matter experts were produced. The mean Intersection over Union and micro F-score metrics were compared after training for 50,000 epochs. This work is illustrative for researchers setting out to develop deep learning models for detection and location of specialist features.

\end{abstract}

\section{Introduction}

	Corrosion is a difficult subject to detect compared to other common subjects such as the human face that have distinct features: two eyes, a nose and a mouth; corrosion shares limited characteristics in colour range and texture - and the appearance of corrosion is confused on both counts, with shadows, boulders, bricks, safety vests all presenting false positives for Deep Convolutional Neural Network (D-CNN) models. Furthermore, the boundary between corroded and uncorroded areas in images is often undefined, due to image compression artefacts and focal range (depth of field).

To produce a useful detector requires semantic segmentation of images rather than simply image classification, i.e., per pixel labelling rather than image level labelling. To achieve this a Fully Convolutional Network (FCN) \cite{Long2015} was employed using strongly supervised training; requiring a dataset of images labelled as densely as the desired output. Ideally, the model can incorporate expert level decision making regarding the severity of corrosion detected. Creating a dataset for Deep Learning of corrosion segmentation requires subject knowledge from labellers to avoid mislabelling of ground truths. The current research is investigating the efficacy of D-CNN for automated detection of corrosion in rapid infrastructure inspections. Within the present paper the balance between a large dataset with poorly labelled ground-truths and a small dataset with expertly annotated segmentations is investigated.

\subsection{Related Work}
	Impressive progress in D-CNN for image classification tasks have been driven in large part by the availability of massive labelled datasets such as ImageNet \cite{JiaDeng2009}. Within the field of semantic segmentation, where each pixel is assigned a class prediction by the model, the largest public datasets include PASCAL VOC \cite{Everingham2010} and MS-COCO \cite{Lin2014}, labelled via a supervised annotation event for the former, and leveraging the Amazon Mechanical Turk (with a specialised user interface) for the latter. These large datasets have an advantage in that the subject matter is easily recognizable by the general public, MS-COCO claims the objects are easily recognizable by a 4-year old. When it comes to specialised subject matter labelling the dataset becomes much more challenging - for example the BRATS dataset of brain tumours \cite{Menze2016} was produced from just 65 MRI scans annotated by seven expert radiographers, with each segmentation taking approximately 60 minutes.	
    
	The difficulty in producing semantic segmentation datasets is recognised and tackled by many researchers using novel approaches to automate dataset creation. The simple-to-complex approach \cite{Wei2015} uses saliency mapping from image classification to generate a dataset for training a simple semantic segmentation model, that then produces a dataset for a more complex semantic segmentation model. \cite{Bearman2016} use point clicks on subject objects as a semi-automated method to generate rough semantic segmentation maps. Otherwise fine-tuning of pre-trained networks on specialist subjects has proven successful \cite{Milan2017}, especially where the specialist dataset shares common lower level features with the pre-trained dataset.
    
	Mislabelled data was found to negatively impact classification performance more than if it was excluded from the dataset by \cite{Reale2016} - with just 10\% mislabelled data roughly equivalent to halving the dataset size. However, \cite{Rolnick2017} found that larger model architectures are able to deal with significant label noise provided it is not adversarial, achieving good performance with less than 1\% better than chance on classification labels - although a maximum useful size of dataset was found, beyond which performance improvements were marginal. Intuitively, larger neural network architectures are able to learn more representations of features, which provides an increase in robustness to noisy training signals. The relationship between dataset size and classification accuracy has also been investigated by \cite{Cho2015} who found that increasing dataset size increases accuracy, and formulated a method to estimate the dataset size required to achieve target accuracy for a specific task by training on varying dataset size.
    
	Incorrectly labelled semantic segmentation features can be considered adversarial if they are detrimental to learning, i.e., they confuse the model by providing feature level false positives and negatives in the training set. Whereas, non-adversarial noise comprises edge cases where the boundary is not clear, these can be overcome by the sheer number of correctly labelled pixels. To the authors' knowledge no previous work has been undertaken to evaluate the impact of label noise on deep learning semantic segmentation models.

\section{Methodology}
\begin{figure}[h!]
\centering
 \subfloat[Input Image]{\framebox{\includegraphics[width=0.3\textwidth]{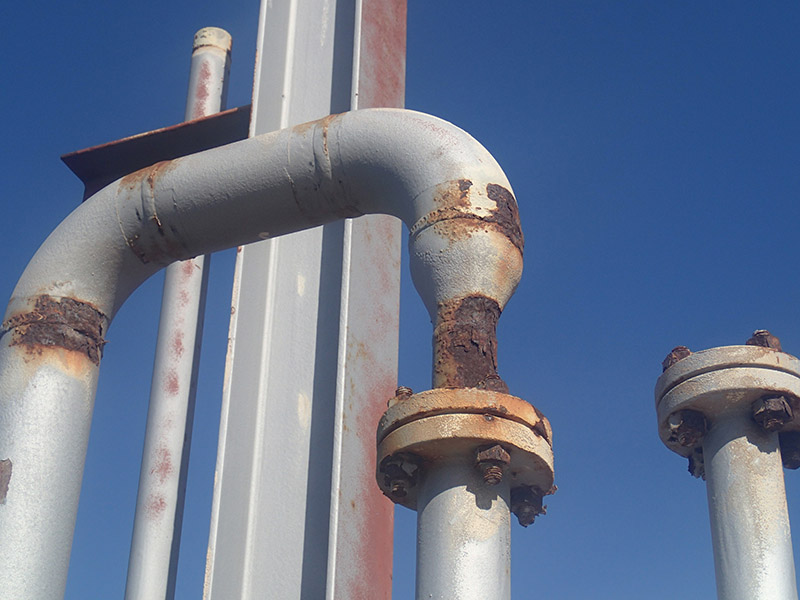}\label{fig:input_Ass-1}}}
 \subfloat[DS-A labels]{\framebox{\includegraphics[width=0.3\textwidth]{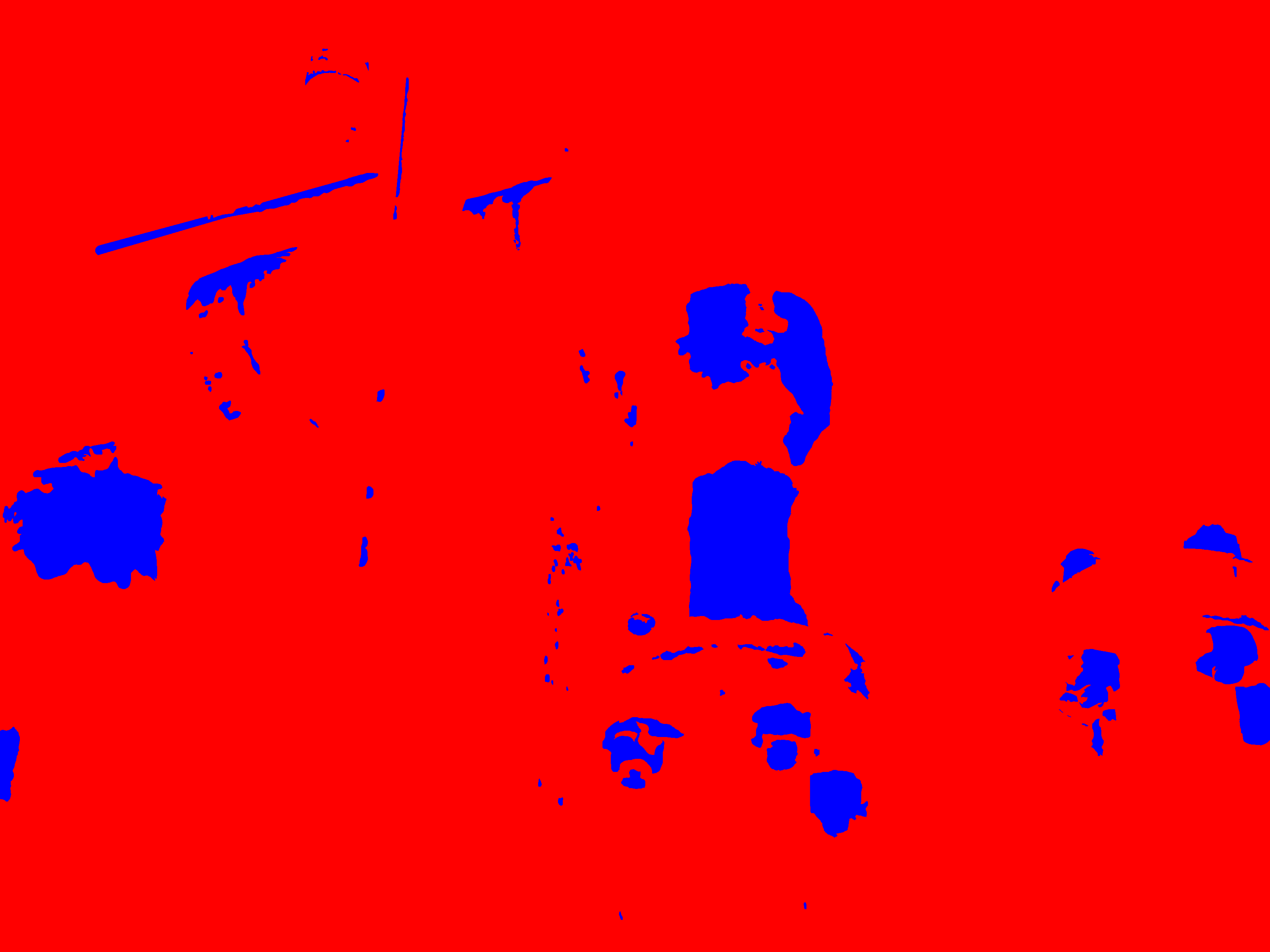}\label{fig:Ass-1_GTa_2}}}
 \subfloat[DS-B labels]{\framebox{\includegraphics[width=0.3\textwidth]{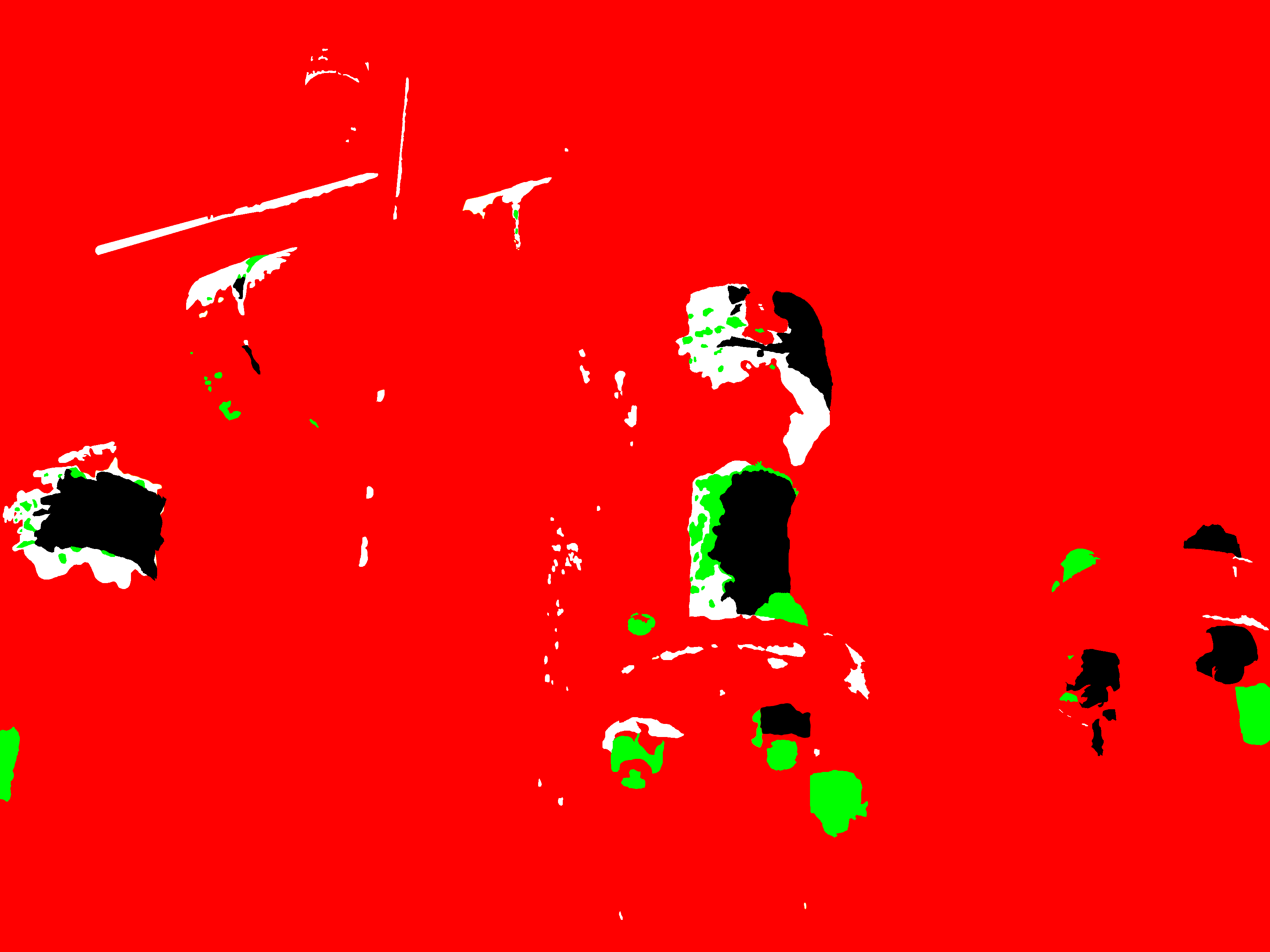}\label{fig:Ass-1_GTb_2}}}
%
%
\caption{Example Assessment Image for comparing DS-A and DS-B models. For DS-A: blue = corrosion class; and red = background class. For DS-B: white = minor corrosion class; green = moderate corrosion class; black = severe corrosion class; and red = background class}
\label{fig:example}
\end{figure}

Two competing Datasets (DS) were produced to test the efficacy of training with a large `imperfect' DS against the performance of training with a small `perfect' DS:
\begin{itemize}
 \item DS-A is used to denote the larger data set of 250 images, imperfectly labelled for two classes: corrosion and background. An 80 / 20 split was used for training and validation sets respectively. 
 \item DS-B refers to the smaller data set of ten images, expertly segmented for corrosion in four classes: minor, moderate and severe corrosion, plus background. Training was undertaken on five images with two images reserved for validation.
\item Three `Assessment' images were reserved and expertly segmented for the performance comparison of the two models.Figure \ref{fig:example} presents an example assessment image and the associated labelled segmentations for the two models.\\
\end{itemize}

The FCN used herein was based on the modified VGG-16 architecture \cite{Simonyan2014a}. Python code was adapted from the TensorFlow implementation of \cite{Teichmann2016}, which was successfully used for semantic segmentation of road scenes from the KITTI autonomous driving dataset \cite{Fritsch2013}. The FCN was trained using transfer learning for 50,000 iterations on an NVIDIA GTX 1080Ti with the following hyper-parameters: Optimizer: Adam, Learning Rate: 10\textsuperscript{-5}, Batch Size: 1, Loss: Cross Entropy.

Performance of the two models was assessed by comparing the confusion matrices across classes; and calculating the mean Intersection over Union (mIoU) and micro F-scores for the three assessment images. To provide a fair comparison the DS-B metrics were computed for `all corrosion', where confusion between the corrosion classes was ignored. The `background' class was excluded from the metrics because it vastly outnumbers the corrosion pixels.

\begin{equation}
\text{mIoU} = \frac{_1}{_N}{\sum_{2}^{N}} \frac{TP}{(TP + FP + FN)} \\
\label{mIoU_eqn}
\end{equation}

\begin{equation}
\text{micro F-score} = \frac{\sum_{2}^{N}2TP}{\sum_{2}^{N}(2TP + FP + FN)} \\
\label{Fscore_eqn}
\end{equation}

Equations \ref{mIoU_eqn} and \ref{Fscore_eqn} present the formula for computing the mIoU and micro F-score respectively; where:
TP = True Positives,
FP = False Positives, and
FN = False Negatives.

%
%
\newpage
\section{Results}

\begin{figure}[!h]
 \begin{minipage}[b]{\textwidth}
 \centering
  \subfloat[][Image 1]{\framebox{\includegraphics[width=0.3\textwidth]{Ass-1.JPG}\label{fig:Ass-1}}}
  \subfloat[][Image 2]{\framebox{\includegraphics[width=0.3\textwidth]{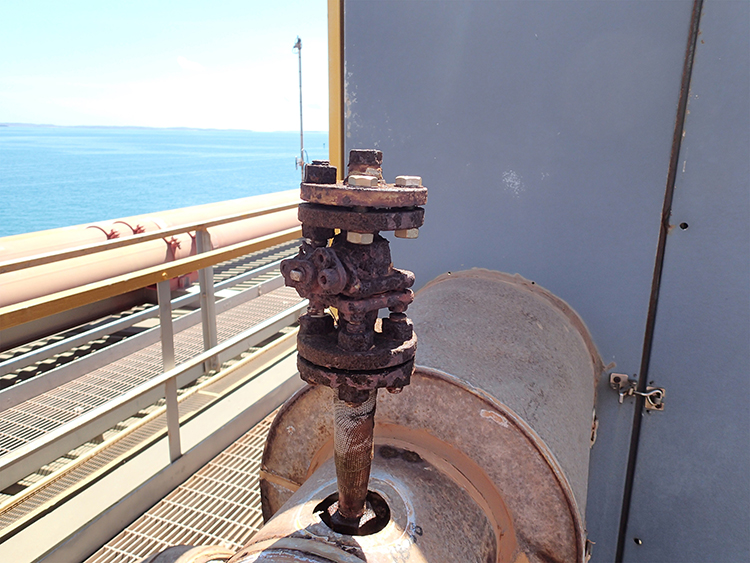}\label{fig:Ass-2}}}
  \subfloat[][Image 3]{\framebox{\includegraphics[width=0.3\textwidth]{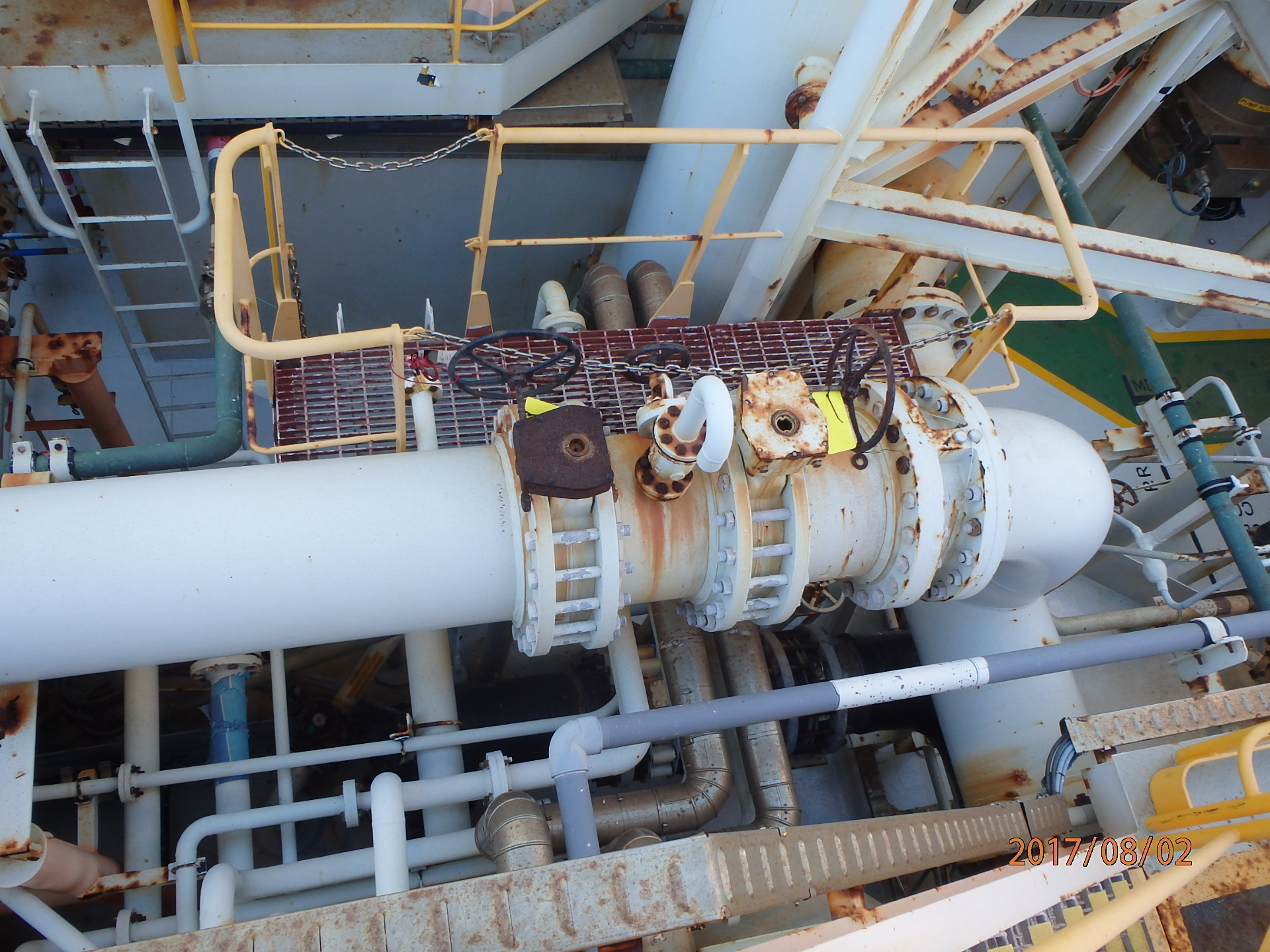}\label{fig:Ass-3}}}
 \end{minipage}
  \begin{minipage}[b]{\textwidth}
 \centering
  \subfloat[][Image 1 DS-A output]{\framebox{\includegraphics[width=0.3\textwidth]{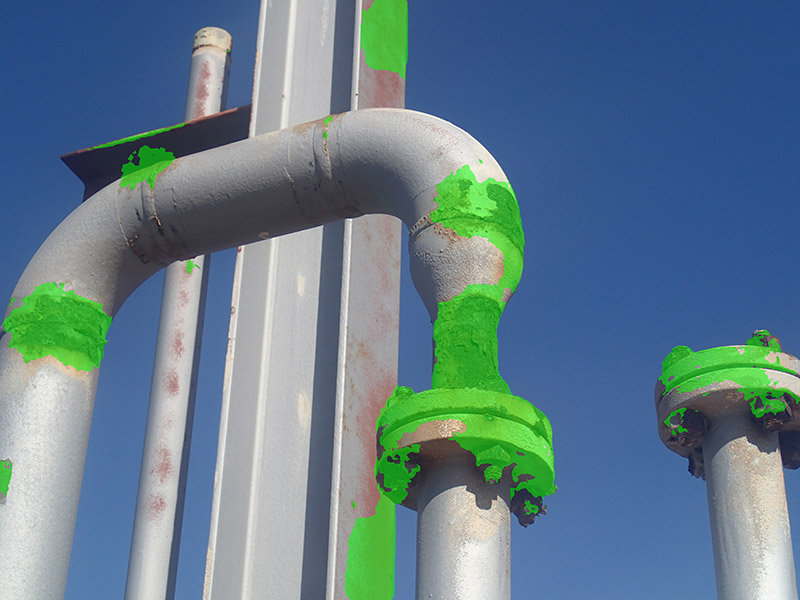}\label{fig:Ass-1_TDS-A_detect}}}
 \subfloat[][Image 2 DS-A output]{\framebox{\includegraphics[width=0.3\textwidth]{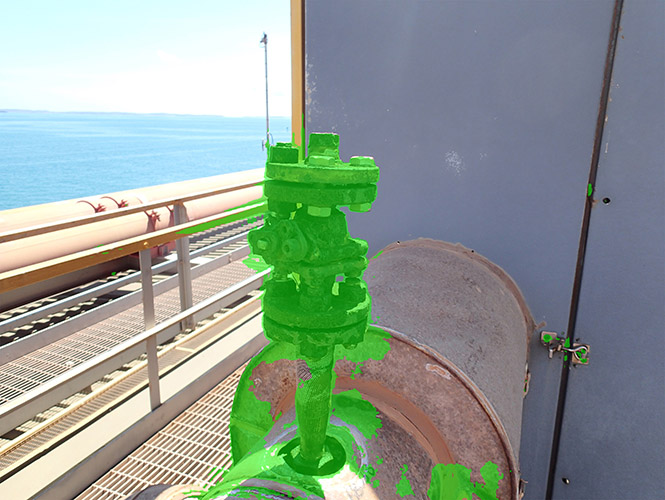}\label{fig:Ass-2_TDS-A_detect}}}
 \subfloat[][Image 3 DS-A output]{\framebox{\includegraphics[width=0.3\textwidth]{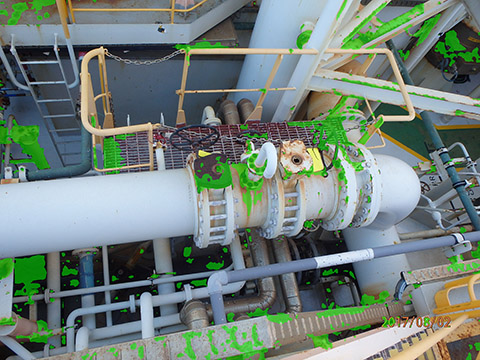}\label{fig:Ass-3_TDS-A_detect}}}
 \end{minipage}
   \begin{minipage}[b]{\textwidth}
 \centering
  \subfloat[Image 1 DS-B output]{\framebox{\includegraphics[width=0.3\textwidth]{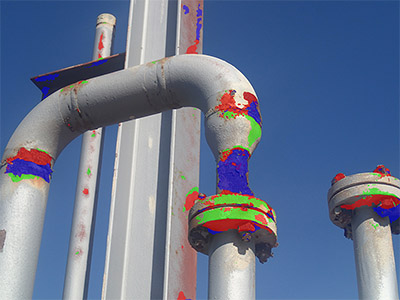}\label{fig:Ass-1_TDS-B_detect}}}
 \subfloat[Image 2 DS-B output]{\framebox{\includegraphics[width=0.3\textwidth]{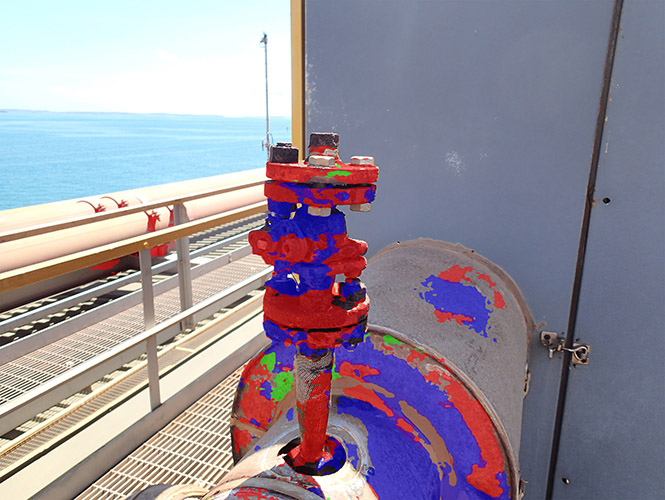}\label{fig:Ass-2_TDS-B_detect}}}
 \subfloat[Image 3 DS-B output]{\framebox{\includegraphics[width=0.3\textwidth]{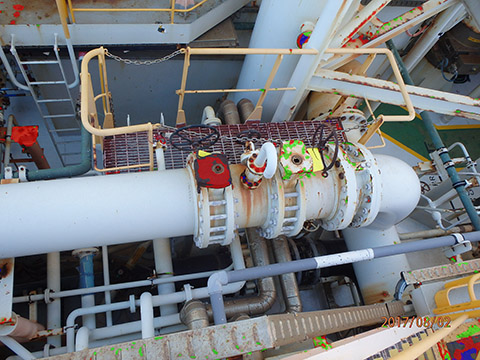}\label{fig:Ass-3_TDS-B_detect}}}
 \end{minipage}
 \caption{\textbf{Output predictions of models.} (a), (b), and (c): assessment images; (d), (e), and (f): DS-A prediction; green = corrosion; (g), (h), and (i): DS-B prediction; green = minor corrosion; red = moderate corrosion; and blue = severe corrosion. }
\label{fig:detection}
\end{figure}

The three assessment images and their respective model predictions are presented in Figure \ref{fig:detection}.

\begin{figure}[h!]
 \begin{minipage}[b]{\textwidth}
 \centering
  \subfloat[DS-A Image 1 Accuracy]{\framebox{\includegraphics[width=0.3\textwidth]{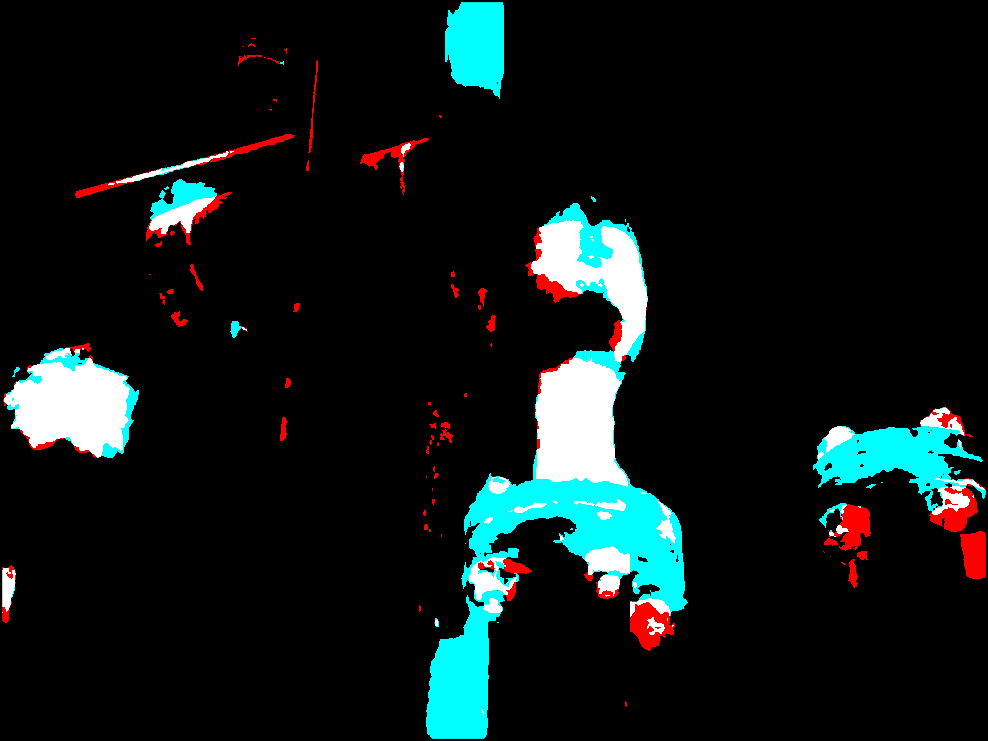}\label{fig:Ass-1_TDS-A}}}
 \subfloat[DS-A Image 2 Accuracy]{\framebox{\includegraphics[width=0.3\textwidth]{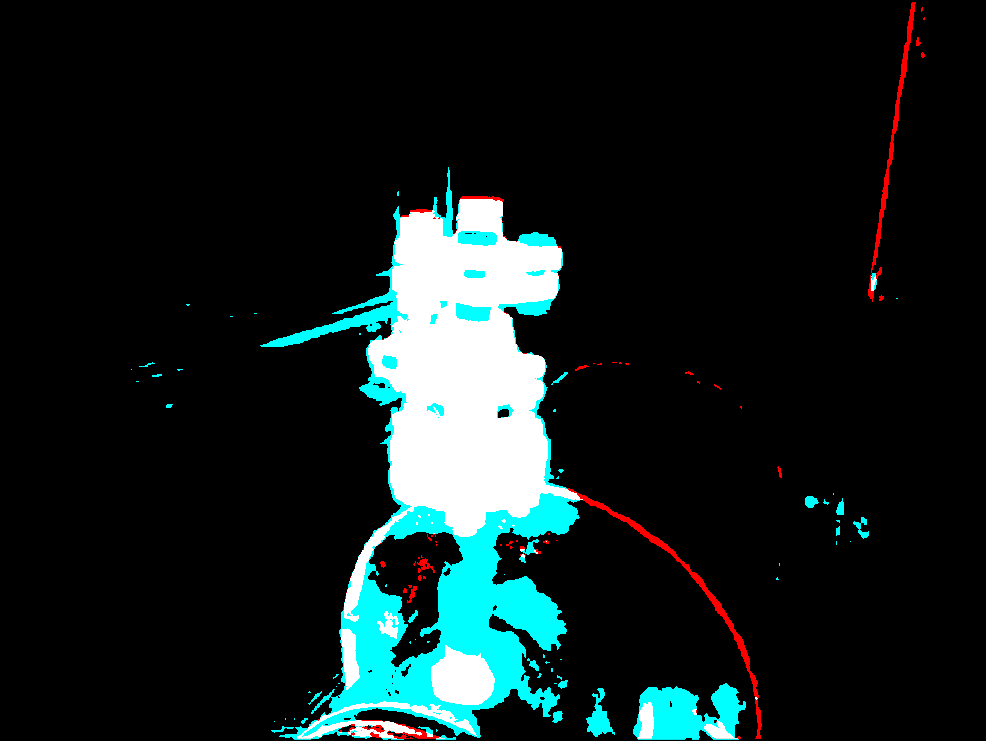}\label{fig:Ass-2_TDS-A}}}
 \subfloat[DS-A Image 3 Accuracy]{\framebox{\includegraphics[width=0.3\textwidth]{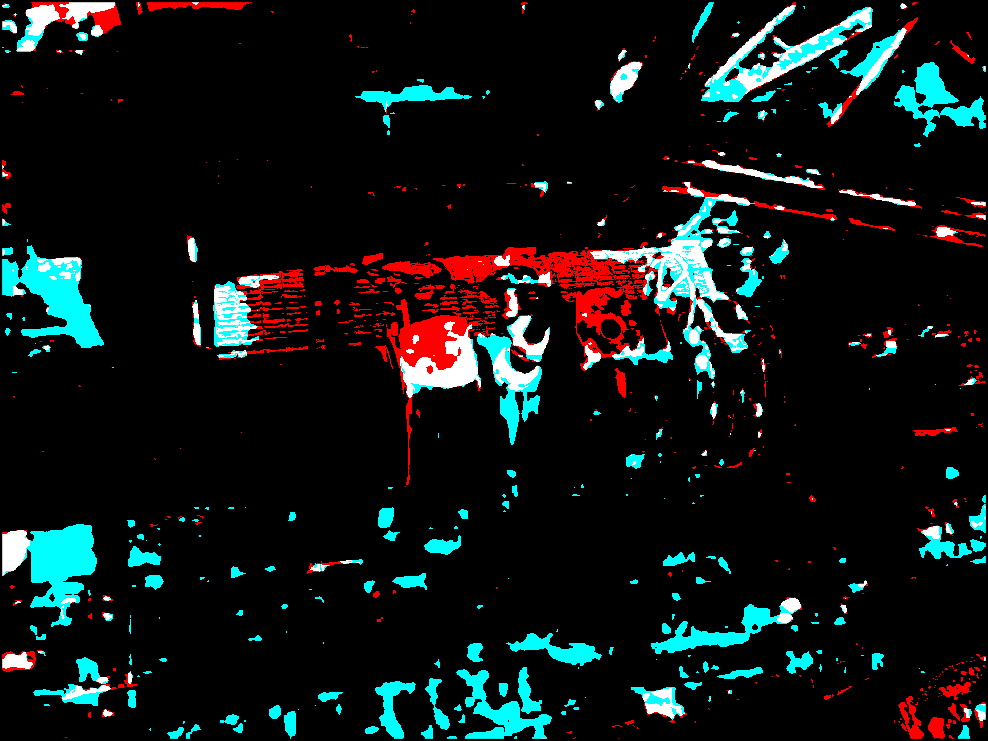}\label{fig:Ass-3_TDS-A}}}
 \end{minipage}
 \begin{minipage}[b]{\textwidth}
 \centering
  \subfloat[DS-B Image 1 Accuracy]{\framebox{\includegraphics[width=0.3\textwidth]{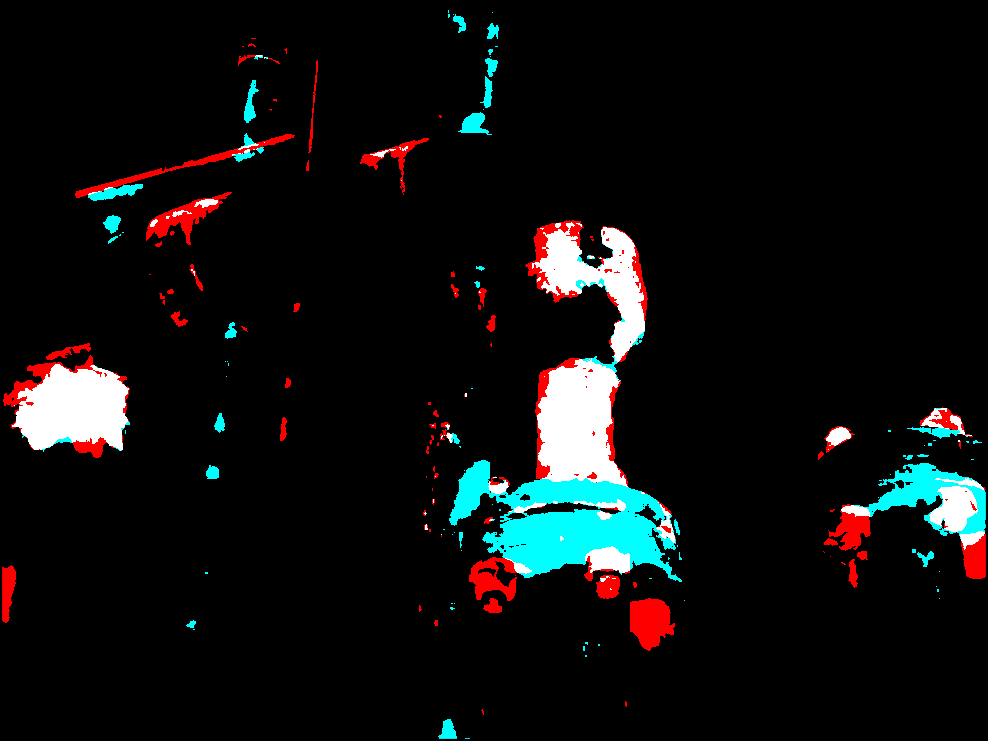}\label{fig:Ass-1_TDS-B}}}
 \subfloat[DS-B Image 2 Accuracy]{\framebox{\includegraphics[width=0.3\textwidth]{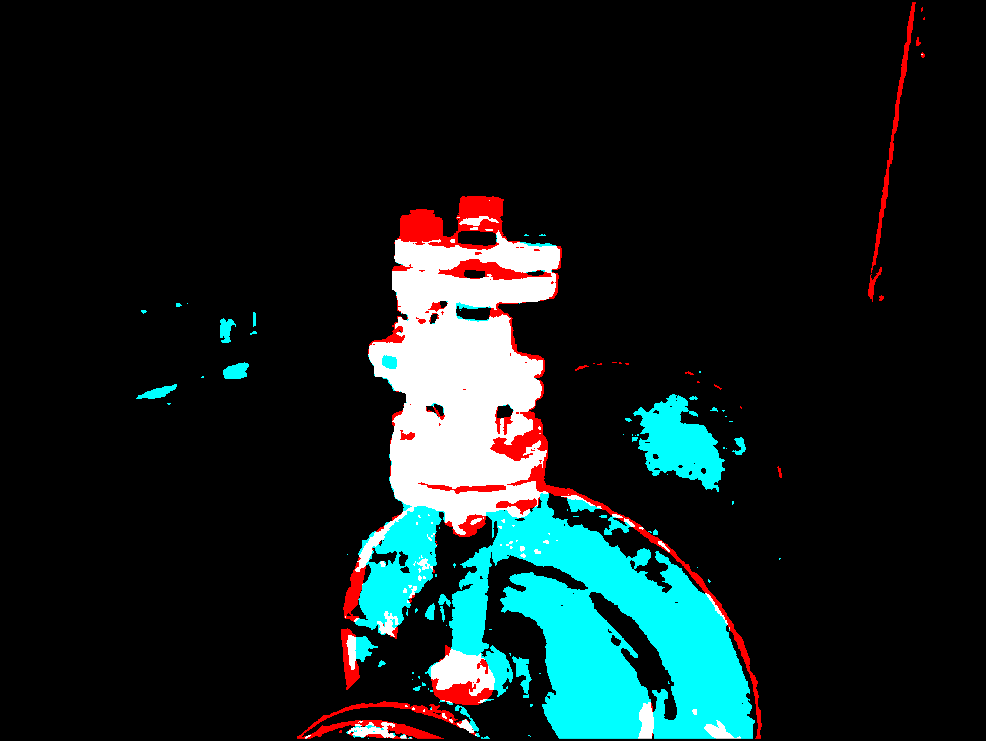}\label{fig:Ass-2_TDS-B}}}
 \subfloat[DS-B Image 3 Accuracy]{\framebox{\includegraphics[width=0.3\textwidth]{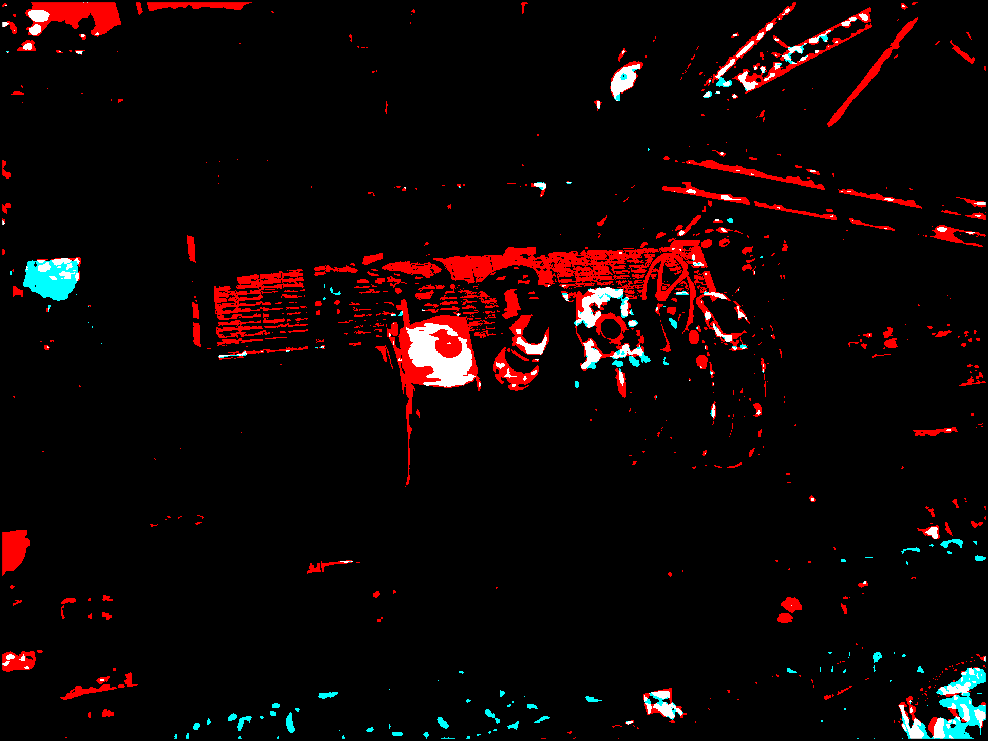}\label{fig:Ass-3_TDS-B}}}
 \end{minipage}
\caption{\textbf{Comparison of DS-A and DS-B models accuracy.} (a), (b), and (c) show the accuracy of the DS-A model; (d), (e), and (f) show the accuracy of the DS-B model. White = True Positive, black = True Negative, red = False Negative, cyan = False Positive}
\label{fig:accuracy}
\end{figure}

\newpage
Figure \ref{fig:accuracy} presents the accuracy of the output predictions from the models compared to the expertly segmented labels. The corrosion prediction accuracy was generated by multiplying the background class segmentation in red by the background class prediction in cyan.

The confusion matrices for the DS-A and DS-B model outputs of the three assessment images are provided in Tables \ref{tab:DS-A_cm} and \ref{tab:DS-B_cm} respectively. For the DS-B model the per class metrics and combined `all corrosion' class metrics (in bold) are presented.

\begin{table}[h!]
\centering
\caption{\textbf{DS-A Confusion Matrix:} Metrics to assess accuracy performance of the DS-A model, including mean Intersection over Union (mIoU) and F-score}
\label{tab:DS-A_cm}
\begin{tabular}{|c|l|r|r|c|r|r|}
\hline
\multicolumn{1}{|l|}{\multirow{2}{*}{Image}} & \multirow{2}{*}{Prediction} & \multicolumn{2}{c|}{Label} & \multicolumn{1}{l|}{\multirow{2}{*}{mIoU}} & {\multirow{2}{*}{F-score}} \\ \cline{3-4} 
\multicolumn{1}{|l|}{}&&\multicolumn{1}{l|}{background} & \multicolumn{1}{l|}{corrosion} & \multicolumn{1}{l|}{}  & \multicolumn{1}{l|}{}\\ \hline
\multirow{2}{*}{1} & background & 4,768,771 & 251,161 & \multirow{2}{*}{0.43} & \multirow{2}{*}{0.60} \\ \cline{2-4} & corrosion & 62,883 & 237,235   &  & \\ \hline
\multirow{2}{*}{2} & background & 4,654,725 & 295,996   & \multirow{2}{*}{0.53} & \multirow{2}{*}{0.69} \\ \cline{2-4} & corrosion & 18,857 & 350,422  &  &  \\ \hline
\multirow{2}{*}{3} & background & 4,603,629 & 431,649 & \multirow{2}{*}{0.22} & \multirow{2}{*}{0.36} \\ \cline{2-4}
& corrosion & 124,488 & 160,234 &  &  \\ \hline
\multicolumn{4}{|r|}{\textbf{Average}} & \textbf{0.39} & \textbf{0.55} \\ \hline
\end{tabular}
\end{table}

\begin{table}[!h]
\centering
\caption{\textbf{DS-B Confusion Matrix:} Compilation of metrics to assess accuracy performance of the DS-B model, including mean Intersection over Union (mIoU) and F-score. The multiple classes of corrosion (minimum, moderate and severe) were combined into `all corrosion' (bolded) to provide a fair comparison with the DS-A model.}
\label{tab:DS-B_cm}
\begin{tabular}{|c|c|r|r|r|r|r|r|r|}
\hline
\multirow{2}{*}{Image} & \multirow{2}{*}{Prediction} & \multicolumn{5}{c|}{Label} & \multirow{2}{*}{mIoU} & \multirow{2}{*}{F-score} \\ \cline{3-7}
& & \multicolumn{1}{c|}{background} & \multicolumn{1}{c|}{min.} & \multicolumn{1}{c|}{mod.} & \multicolumn{1}{c|}{sev.} & \multicolumn{1}{c|}{all corr'n} & & \\ \hline
\multirow{5}{*}{1} & background & \textbf{4,818,964} & 174,523 & 32,033 & 7,230 & \textbf{213,786} & & \\ \cline{2-9} 
 & min. & 48,021 & 25,076 &  22,784 &  9,617 & & 0.09 & 0.16 \\ \cline{2-6} \cline{8-9}  
 & mod. & 33,451 & 735 & 11,077 & 9,668 &- & 0.07 & 0.13 \\ \cline{2-6} \cline{8-9}  
 & sev. & 294,408 & 1,252 & 50,288 & 45,873 & & 0.11 & 0.20 \\ \cline{2-9} 
 & all corr'n & \textbf{379,880} &\multicolumn{3}{|c|}{-} & \textbf{176,370} & \textbf{0.23} & \textbf{0.17} \\ \hline
\multirow{5}{*}{2} & background & \textbf{4,839,149} & 9,027 & 74,048 & 43,601 & \textbf{126,676} & & \\ \cline{2-9} 
 & min. & 16,248 & 456 & 28,170 & 6,561 & & 0.01 & 0.01 \\ \cline{2-6} \cline{8-9}  
 & mod. & 73,221 & 1,494 & 56,621 & 7,916 &- & 0.20 & 0.33 \\ \cline{2-6} \cline{8-9}  
 & sev. & 95,918 & 1,588 & 46,699 & 19,283 & & 0.09 & 0.16 \\ \cline{2-9} 
 & all corr'n & \textbf{185,387} &\multicolumn{3}{|c|}{-} & \textbf{168,788}  & \textbf{0.35} & \textbf{0.26} \\ \hline
\multirow{5}{*}{3} & background & \textbf{5,017,894} & 31,734 & 20,993 & 6,130 & \textbf{58,857}  &- & \\ \cline{2-9} 
 & min. & 82,131 & 47,002 & 20,365 & 5,604 &  & 0.25 & 0.40 \\ \cline{2-6} \cline{8-9} 
 & mod. & 76,959 & 2,715 & 4,076 & 2,933 &- & 0.03 & 0.06 \\ \cline{2-6} \cline{8-9}
 & sev. & 236 & 224 & 624 & 380 & & 0.02 & 0.05 \\ \cline{2-9} 
 & all corr'n & \textbf{159,326} & \multicolumn{3}{|c|}{-} &\textbf{83,923}  & \textbf{0.28} & \textbf{0.27} \\ \hline
\multicolumn{7}{|r|}{\textbf{Average}}  & \textbf{0.29} & \textbf{0.23} \\ \hline
\end{tabular}
\end{table} 
\bigskip

\begin{figure}[!h]
 \begin{minipage}[b]{\textwidth}
 \centering
  \subfloat[Labels]{\framebox{\includegraphics[width=0.3\textwidth]{Ass-1_GTb.png}\label{fig:Ass-1_GTb}}}
  \subfloat[Output]{\framebox{\includegraphics[width=0.3\textwidth]{Ass-1_GTb_detect.jpg}\label{fig:TDS-b_pred}}}
    \subfloat[Background]{\framebox{\includegraphics[width=0.3\textwidth]{TDS-B_bg.png}\label{fig:TDS-B_bg}}}
 \end{minipage}
 \begin{minipage}[b]{\textwidth}
 \centering

  \subfloat[Minor corr'n]{\framebox{\includegraphics[width=0.3\textwidth]{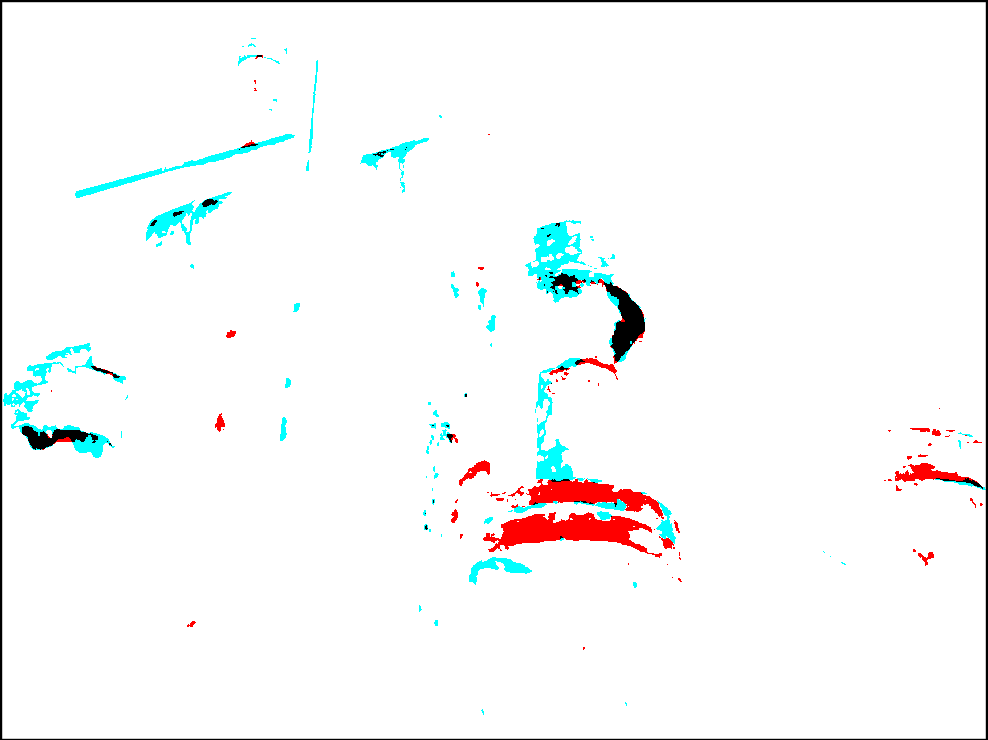}\label{fig:TDS-B_min}}}
    \subfloat[Moderate corr'n]{\framebox{\includegraphics[width=0.3\textwidth]{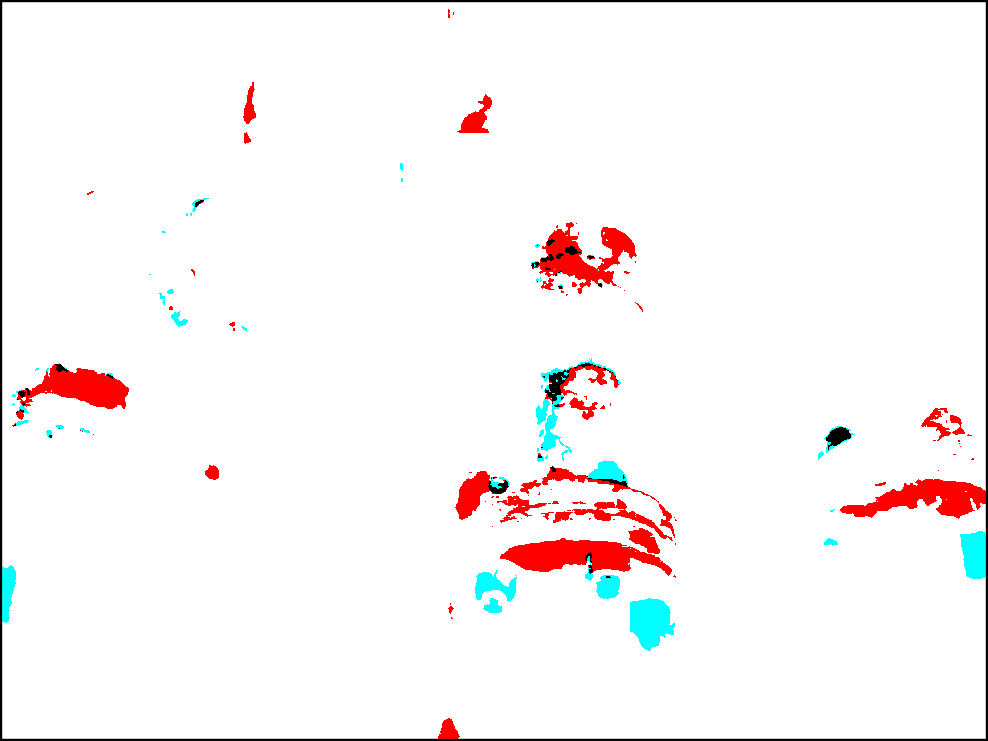}\label{fig:TDS-B_mod}}}
  \subfloat[Severe corr'n]{\framebox{\includegraphics[width=0.3\textwidth]{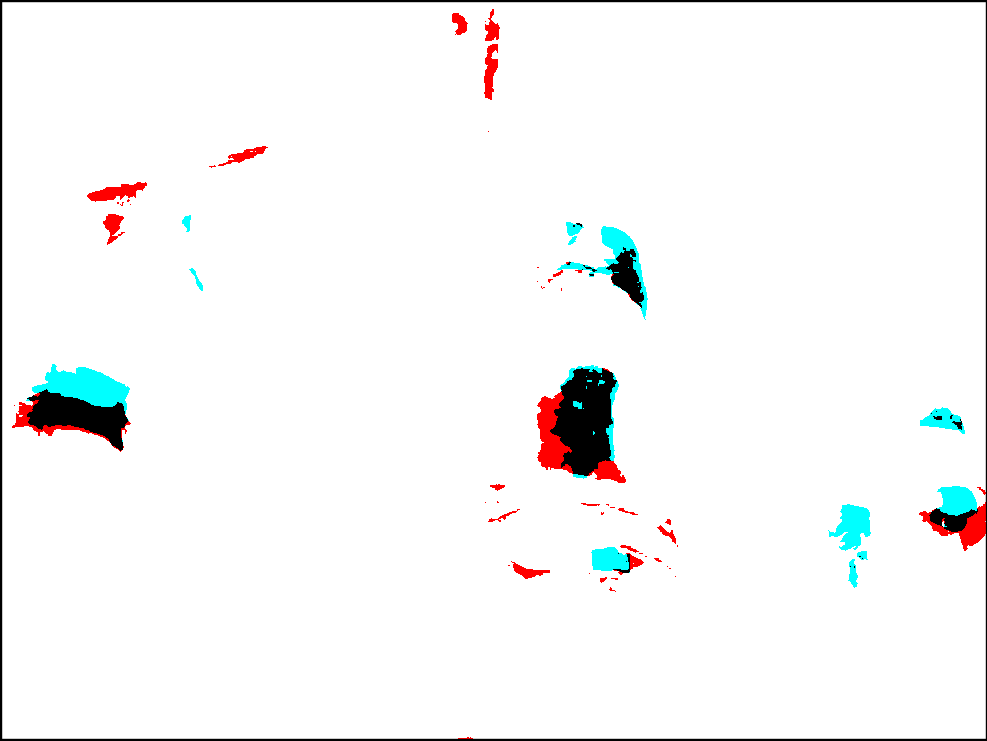}\label{fig:TDS-B_sev}}}
 \end{minipage}

\caption{\textbf{Performance of DS-B multi-class model.} (a) expert segmentation: white = minor corrosion; green = moderate corrosion; black = severe corrosion; and red = background. (b) prediction: green = minor corrosion; red = moderate corrosion; and blue = severe corrosion. (c) to (f) individual class accuracy: black = True Positive, white = True Negative, cyan = False Negative, red = False Positive.}
\label{fig:DS-B_results}
\end{figure}

An example output and the accuracy of the multi-class model trained on DS-B is presented below (Figure \ref{fig:DS-B_results}). Again, the accuracy images were generated by multiplying the label segmentation in red with the prediction segmentation in cyan.

\clearpage
\section{Discussion}

	The first observation to be made is that neither model achieves human level accuracy for detection of corrosion. This demonstrates the difficulty of corrosion as a subject of image detection. Several methods are available to improve the accuracy of detection, such as multi-task learning \cite{Dharmasiri2017} to help the model contextualize the information. 
    
	Generally, the micro F-scores indicate that training with a large, noisy dataset is better for semantic segmentation than training on a small expertly labelled dataset. Implicitly, the noise from the DS-A dataset is overcome by having more data for training, whereas limited data available from DS-B is insufficient to train to accurate levels.
    
    The performance drops considerably for the DS-B model on Image 3. While similar training images are included in DS-A, DS-B has no such images indicating that the DS-B strategy suffers from over-fitting. The issue of over-fitting on small datasets is well recognised and documented in \cite{Taigman2014} for faces, \cite{Lu2016} for medical imaging, and an excellent discussion of the issue of over-fitting can be found in \cite{Babyak2004}. Recent work by \cite{Shwartz-Ziv2017} indicates that smaller datasets are prone to over-fitting as the model compresses and discards extraneous information in the network. Although the expert labelling provides more information, DS-B provides a total of only 27 million pixels, and these pixels do not provide independent information, therefore over-fitting with 134 million parameters is not surprising - it should be noted that for this particular application over-fitting is not considered to be significantly detrimental to the intended end use.
    
    Finally, the practical approach to achieving multi-class segmentation with sufficient accuracy would seem to be training the network on a large `imperfect' dataset, and then fine-tuning on a small, expertly segmented, dataset.

\section{Conclusions}

The work presented herein demonstrates that training with a larger, imperfectly segmented dataset outperforms a very small, expertly segmented dataset. Intuitively the small dataset doesn't provide sufficient examples for the model to learn a general representation of the subject; consequently the model suffers both low accuracy and over-fitting. Furthermore, the larger dataset provides sufficient number of accurate segmentations to overcome the noise. Therefore, for specialist subject matters it is preferable to build a large dataset at the expense of introducing noise to the segmentations. Finally, it is suggested that, in the context of corrosion detection, a viable strategy would be to first train on the large dataset, before fine-tuning on the expert labelled dataset to both improve accuracy and increase discrimination of discrete classes.



\newpage


\section{Acknowledgement}

The authors wish to acknowledge the support provided by Woodside Petroleum in providing images of corrosion.

\section{Author Contributions}

WN prepared the datasets, performed training and analysis of the models, and prepared the manuscript (50\%), TD provided technical guidance for the coding and reviewed the manuscript (20\%) NB provided technical support for expert labelling of images and reviewed the manuscript (30\%), 

\section{Conflict of Interest}

To the authors' knowledge there are no conflicts of interest to declare.


\newpage

\appendix









\end{document}